\documentclass{article}
\usepackage[preprint]{neurips_2025}

\usepackage[utf8]{inputenc}
\usepackage[T1]{fontenc}
\usepackage{hyperref}
\usepackage{url}
\usepackage{booktabs}
\usepackage{amsfonts}
\usepackage{amsmath}
\usepackage{nicefrac}
\usepackage{microtype}
\usepackage{xcolor}
\usepackage{graphicx}
\usepackage{subcaption}
\usepackage{wrapfig}

\title{Attention Manifolds: Steering Language Models \\ by Editing Learned B-Spline Surfaces}

\author{
  Naveen Mysore \\
  University of California, Santa Barbara \\
  Dyssonance AI \\
  \texttt{nmysore@ucsb.edu} \\
  \texttt{nmysore.work@gmail.com}
}

\begin{document}

\maketitle

\begin{abstract}
In standard transformer attention, a source token sends the same value vector to every receiver. The query determines \emph{how much} to attend but not \emph{what} to extract. This work introduces \textbf{attention manifolds}: learned 2D B-spline surfaces $S_d(q_d, k_d)$ that modulate each value dimension based on the query-key interaction. Each surface is a tensor-product cubic B-spline initialized to zero, preserving pretrained behavior. Applied to LLaMA 3.2-1B-Instruct and 3B-Instruct, attention manifolds reduce WikiText-2 validation perplexity by 2--2.5 points with 0.3\% parameter overhead. Across 112 diverse prompts, surfaces change greedy-decoded output for 69\% (1B) to 83\% (3B) of cases, with the strongest effects on ambiguous and polysemous inputs (94--100\% change rate). The surfaces improve output quality: correcting factual errors (\emph{``the CAP theorem has three main components''} $\to$ \emph{``it is impossible to guarantee all three''}), increasing precision (\emph{``impossible to know certain properties''} $\to$ \emph{``impossible to know both position and momentum''}), and adding specificity (a generic quote $\to$ an attributed Saint Augustine citation, consistently at both scales). The learned surfaces are also mechanically editable: inverting a layer's coefficients changes greedy output for 9/10 prompts (KL~0.010), providing a geometric mechanism for model steering. Setting surface coefficients to $-1$ creates ``attention walls'' that block value flow through specific dimensions. In a preliminary experiment, a layer-wide wall redirects an explosive-device prompt from specific instructions to general educational content, suggesting a path toward safety-oriented manifold shaping.
\end{abstract}

\section{Introduction}

\begin{wrapfigure}{r}{0.35\textwidth}
\vspace{-20pt}
\centering
\includegraphics[width=0.33\textwidth]{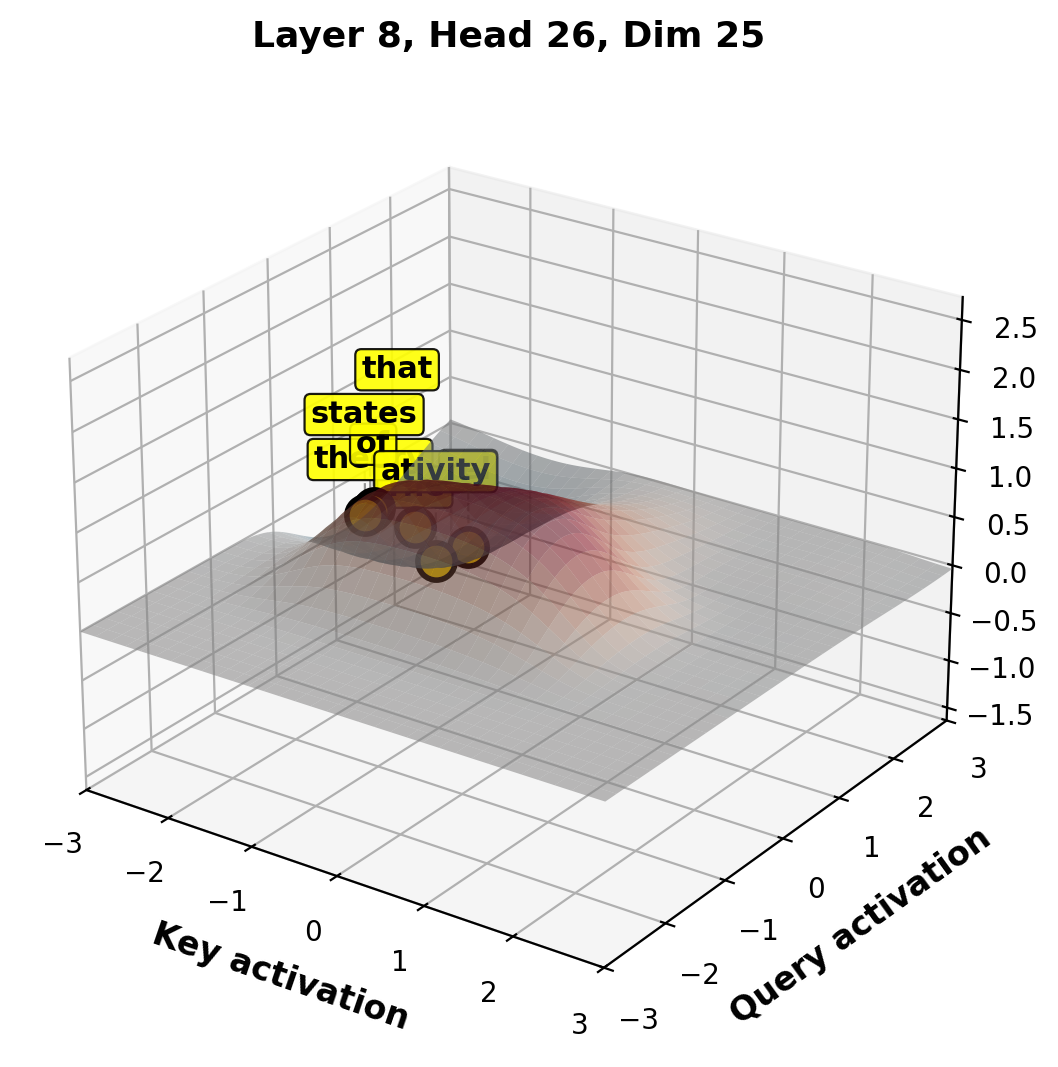}
\caption{Learned surface $S_d(q,k)$. Height = value modulation gain.}
\label{fig:terrain3d}
\vspace{-15pt}
\end{wrapfigure}

Transformer attention~\citep{vaswani2017attention} computes a weighted sum of value vectors, where the weights depend on query-key similarity. The value vector $\mathbf{v}_j$ is identical regardless of which token attends to it: the query controls \emph{how much} but not \emph{what}.

This work adds a per-dimension modulation. For each dimension $d$ of each attention head, a 2D B-spline surface $S_d(q_d, k_d)$ scales the value:
\begin{equation}
o_{id} = \sum_j a_{ij} \cdot v_{jd} \cdot \left(1 + S_d(q_{id}, k_{jd})\right)
\label{eq:value_mod}
\end{equation}
where $S_d(q, k) = \mathbf{b}(q)^\top \mathbf{C}_d \, \mathbf{b}(k)$ is a tensor-product cubic B-spline with learned coefficient matrix $\mathbf{C}_d$, initialized to zero. The model starts as standard attention and discovers improvements during training.

Beyond perplexity, the surfaces improve output quality across diverse prompts: correcting technical errors, increasing factual precision, and resolving ambiguity differently. On polysemous inputs like \emph{``The chicken is ready to eat''} or \emph{``She saw the man with the telescope,''} surfaces change greedy output 94--100\% of the time. The learned surfaces are also \emph{mechanically editable}: inverting a layer's coefficients changes greedy output for 9/10 prompts, providing a geometric mechanism for model steering.

Contributions: (1)~value modulation via zero-initialized B-spline surfaces improving perplexity by 2--2.5 points at 0.3\% overhead on LLaMA 3.2-1B and 3B; (2)~systematic evaluation across 112 prompts showing surfaces improve output quality, especially on ambiguous inputs; (3)~controlled steering experiments showing layer-wide edits reliably alter output, and attention walls that can suppress specific content.

\section{Related Work}
\label{sec:related}

\paragraph{Attention mechanisms.} Scaled dot-product attention~\citep{vaswani2017attention} and its variants (grouped-query attention~\citep{ainslie2023gqa}, linear attention~\citep{katharopoulos2020transformers}) preserve the property that value vectors are query-independent. KArAt~\citep{karat2025} replaces the attention score function with learnable KAN activations for vision transformers but does not modulate values. DiSAN~\citep{shen2018disan} applies feature-wise attention weights, and Nested Attention~\citep{patashnik2025nested} constructs query-dependent attention values for diffusion-model personalization. Gated Attention~\citep{qiu2025gated} adds query-dependent output gating. This work is distinguished by its specific parameterization: a coordinatewise tensor-product B-spline surface~\citep{piegl1997nurbs} conditioned on \emph{both} query and key, applied per dimension to frozen pretrained models, with the resulting control net directly editable.

\paragraph{Splines in neural networks.} KANs~\citep{liu2024kan} replace fixed activations with learnable B-spline functions per edge. InKAN~\citep{inkan2026} provides an efficient implementation via truncated power form. This work uses 2D tensor-product B-spline surfaces not as activation functions but as modulation functions within attention, operating on post-RoPE~\citep{su2024roformer} coordinates.

\paragraph{Model editing and steering.} Representation engineering~\citep{zou2023representation} identifies behavioral directions in activation space and applies direct activation interventions. Contrastive Activation Addition~\citep{rimsky2024steering} steers language model behavior by adding activation differences at inference time. LoRA~\citep{hu2022lora} adds low-rank weight updates for fine-tuning. Attention manifolds differ by providing an explicit, smooth, per-dimension modulation function that is directly visualizable as a 2D surface and editable at the level of individual B-spline control points. The edit is geometric (reshaping a terrain) rather than algebraic (modifying a weight matrix or adding an activation vector).

\section{Method}
\label{sec:method}

\subsection{Value Modulation}

Standard multi-head attention computes $\mathbf{o}_i = \sum_j a_{ij}\,\mathbf{v}_j$ where $a_{ij} = \mathrm{softmax}_j(\mathbf{q}_i^\top\mathbf{k}_j / \sqrt{d})$. Value modulation replaces this with Eq.~\ref{eq:value_mod}, where each surface $S_d$ is parameterized by a $K \times K$ coefficient matrix ($K = \text{grid\_size} + 3$). With grid size~8, each surface has $11 \times 11 = 121$ coefficients.

All coefficient matrices are initialized to zero, ensuring $S_d \equiv 0$ at initialization. No pretrained behavior is lost. Only the surface coefficients are trained; all original model parameters remain frozen. The total overhead is 0.3\% of model parameters (see Appendix~\ref{app:details}).

\subsection{Efficient Computation}

Constructing the modulated tensor $m_{ijd}$ of shape $(\text{batch}, \text{heads}, \text{seq}_q, \text{seq}_k, \text{dim})$ is prohibitive. The computation factors as:
\begin{equation}
\Delta o_{id} = \sum_b \underbrace{\left[\sum_a C_{dab}\,B_a(q_{id})\right]}_{\text{query features}} \cdot \underbrace{\left[\sum_j a_{ij}\,v_{jd}\,B_b(k_{jd})\right]}_{\text{attn-weighted KV features}}
\label{eq:factored}
\end{equation}
The full output is $\mathbf{o} = \mathbf{A}\mathbf{V} + \Delta\mathbf{o}$, avoiding the per-pair tensor while preserving the exact computation.

\section{Experiments}
\label{sec:experiments}

\paragraph{Setup.} Base models are LLaMA 3.2-1B-Instruct and LLaMA 3.2-3B-Instruct~\citep{llama3}. All pretrained parameters are frozen. Surface coefficients are trained with next-token cross-entropy on a 1.28M-token subset of WikiText-103~\citep{merity2017pointer} (see Appendix~\ref{app:details}). Perplexity is evaluated on the full WikiText-2 validation set (252K tokens, strided sliding window with context 512 and stride 256).

\paragraph{Results.}

\begin{table}[h]
\centering
\caption{Value modulation results on LLaMA 3.2-1B and 3B.}
\label{tab:main}
\begin{tabular}{lccc}
\toprule
Model & Params Added & PPL & $\Delta$ \\
\midrule
LLaMA 3.2-1B (baseline) & --- & 14.30 & --- \\
+ Value modulation (grid=8) & 4.0M (0.3\%) & \textbf{12.31} & \textbf{--1.99} \\
\midrule
LLaMA 3.2-3B (baseline) & --- & 11.73 & --- \\
+ Value modulation (grid=8) & 10.4M (0.3\%) & \textbf{9.24} & \textbf{--2.49} \\
+ Value modulation (grid=16) & 30.9M (1.0\%) & 9.25 & --2.48 \\
\bottomrule
\end{tabular}
\end{table}

Grid size~8 achieves the best result at both scales. Grid~16 uses $3\times$ more parameters but does not improve perplexity; the routing structure is captured by smooth surfaces. The improvement scales with model size: --1.99 for 1B and --2.49 for 3B.

\begin{figure}[t]
\centering
\includegraphics[width=\textwidth]{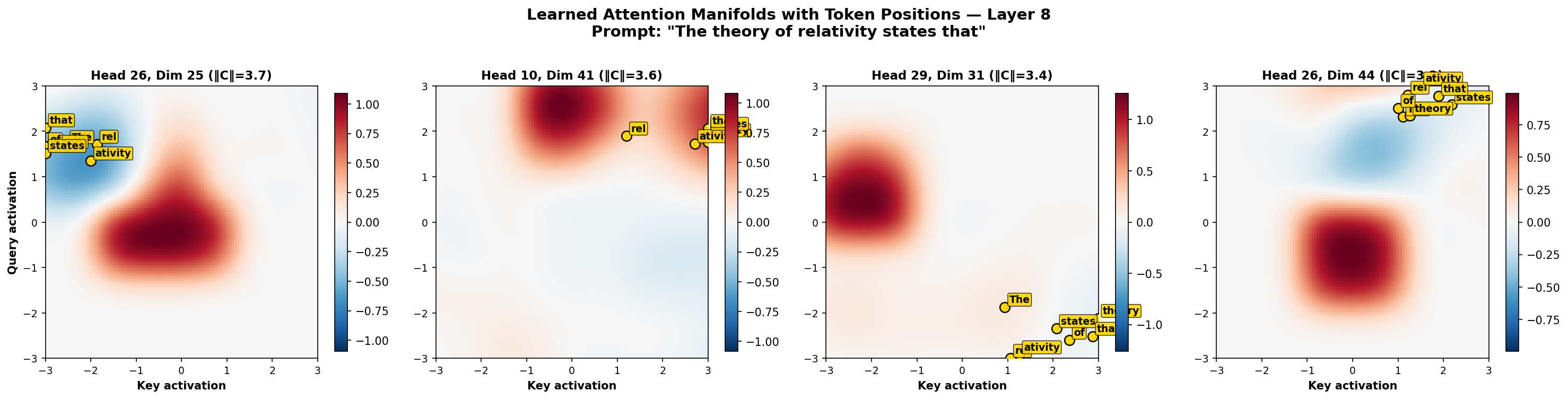}
\caption{Learned attention manifolds for the prompt \emph{``The theory of relativity states that.''}  Each panel shows one surface $S_d(q,k)$. Dots mark each token's self-pair projection $(q_{i,d}, k_{i,d})$; the actual computation evaluates $S_d$ at all receiver--source pairs $(q_{i,d}, k_{j,d})$ weighted by attention. Red regions amplify the value ($S>0$); blue regions attenuate ($S<0$). All coordinates are post-RoPE.}
\label{fig:surfaces}
\end{figure}

\paragraph{Surface geometry.} The learned surfaces exhibit saddle shapes, ridge patterns, and localized bumps (Figure~\ref{fig:surfaces}). Most surfaces per layer have small norms, while a long tail of high-norm surfaces provides strong corrections. Modulation concentrates in specific head-dimension combinations.

\begin{figure}[h]
\centering
\includegraphics[width=\textwidth]{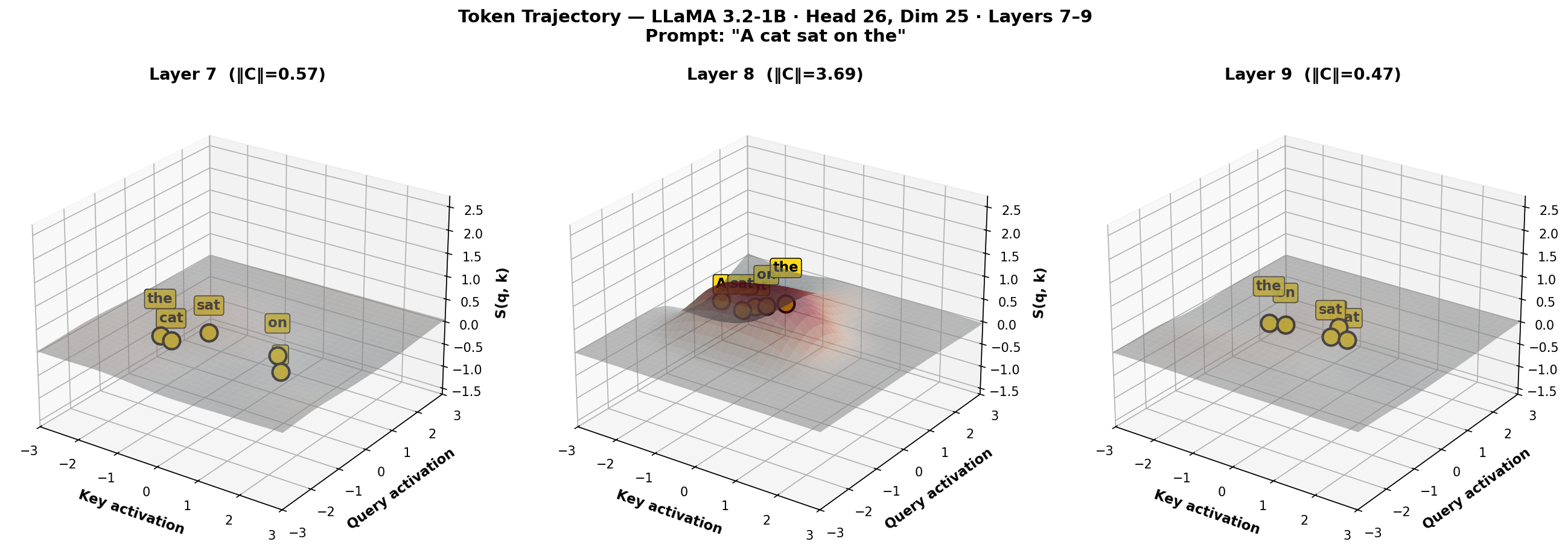}
\caption{Token trajectory across layers 7--9 (Head~26, Dim~25, LLaMA 3.2-1B). Dots show self-pair projections $(q_{i,d}, k_{i,d})$ on post-RoPE coordinates. Layer~8 has the strongest surface ($\|\mathbf{C}\| = 3.69$); adjacent layers are nearly flat.}
\label{fig:trajectory}
\end{figure}

\paragraph{Token trajectories.} As tokens pass through layers, their Q/K activations change, landing at different positions on each layer's surface (Figure~\ref{fig:trajectory}). Each token traces a unique path through the manifold landscape.

\paragraph{Output quality across 112 prompts.} To evaluate beyond perplexity, 112 diverse prompts spanning 12 categories (science, creative, technical, ambiguous, reasoning, etc.) were processed with greedy decoding, comparing baseline and surface-augmented models. Table~\ref{tab:category_summary} in Appendix~\ref{app:examples} reports full results. Key findings: surfaces change greedy output for 69\% (1B) to 83\% (3B) of prompts. Ambiguous and polysemous prompts show the highest change rate (94--100\%) and KL divergence, consistent with the hypothesis that Q-K dependent modulation improves context-dependent routing. Code prompts are least affected (0--40\%), suggesting surfaces primarily modulate semantic rather than syntactic patterns. Representative examples include correcting the CAP theorem answer from ``three main components'' to ``impossible to guarantee all three,'' and sharpening the uncertainty principle from ``certain properties'' to ``both position and momentum with infinite precision'' (see Appendix~\ref{app:examples}).

\section{Manifold Manipulation}
\label{sec:manipulation}

The learned manifolds are mechanically editable. Because each surface is defined by a coefficient matrix $\mathbf{C}_d$, the terrain can be scaled ($\alpha\mathbf{C}_d$), inverted ($-\mathbf{C}_d$, changing the gain from $1+S$ to $1-S$), or zeroed ($\mathbf{0}$). $\mathbf{C}_d$ is an $11 \times 11$ control net (0-indexed): rows are control points along the \emph{query} axis, columns along the \emph{key} axis. The most logit-sensitive layer is identified by ablation (zeroing each layer's surfaces and measuring logit divergence), and the largest-norm surfaces within it are candidates for editing.

We evaluate steering with controlled experiments: greedy decoding (deterministic), paired-seed sampling (10 seeds per condition, temperature~0.8, top-$k$~50), and KL divergence between next-token distributions. All probabilities are single-token under unscaled softmax; ``\texttt{windows}'' is the first token of ``windowsill'' (which tokenizes as two pieces).

\paragraph{Layer-wide inversion.} Inverting all 2,048 surfaces in Layer~8 produces measurable distributional shifts. Table~\ref{tab:controlled} summarizes results across 10 diverse prompts.

\begin{table}[h]
\centering
\small
\caption{Controlled steering: layer-wide vs.\ single-surface inversion (Layer 8). KL divergence measured on next-token distribution. Greedy and paired-seed columns report how many of 10 prompts/seeds change.}
\label{tab:controlled}
\begin{tabular}{lcccc}
\toprule
Intervention & Coefficients & Mean KL & Greedy changed & Seeds changed \\
\midrule
Single surface invert & 121 & 0.0001 & 1/10 & 0--3/10 \\
Single surface zero & 121 & 0.00003 & 0/10 & 0--1/10 \\
Layer invert & 247,808 & 0.010 & 9/10 & 2--7/10 \\
Layer zero & 247,808 & 0.003 & 7/10 & 1--4/10 \\
\bottomrule
\end{tabular}
\end{table}

Layer-wide inversion changes greedy-decoded output for 9 of 10 prompts and alters sampled continuations for 20--70\% of paired seeds. Representative greedy changes include: \emph{``She walked into the old mansion''} $\to$ \emph{``\ldots old, abandoned church''}; \emph{``The river flowed through the village of Kavala''} $\to$ \emph{``\ldots through the valley, its gentle current.''} Single-surface inversion has a small but measurable effect (mean KL = 0.0001), changing greedy output for only 1/10 prompts. The sampled ``throne'' continuation reported in earlier exploratory experiments reflects sampling variability, not a reliable single-surface steering effect.

\paragraph{Baseline variability.} Without any edit, sampling with 10 different seeds produces 10 unique continuations for all tested prompts (first 50 characters), confirming that sampled examples alone cannot establish a steering effect. The controlled experiment above uses paired seeds and greedy decoding to separate intervention effects from sampling noise.

\paragraph{Partial edits to the control net.} Despite the small single-surface KL, the structure of the edit matters. On the one prompt where single-surface greedy output changed, inverting only rows~5--10 (66 of 121 coefficients, covering basis centers $\geq 0$) reproduced the change, while inverting rows~0--4 or key columns did not. Cubic basis functions span 4 knot intervals, so row~5 (centered at~0) also affects negative query coordinates. This demonstrates that row/column edits to the control net provide directional control, even when the overall single-surface effect is small.

\paragraph{Attention walls: blocking unsafe terrain.} If surfaces can steer output by amplifying value dimensions, they can also \emph{suppress} them. Setting $S_d = -1$ at specific Q-K regions makes the gain $1 + S = 0$, completely blocking information flow through that dimension. This creates an ``attention wall'' that prevents the model from routing through certain value channels.

Table~\ref{tab:walls} shows a preliminary experiment: setting $S_d = -1$ across all surfaces in Layer~8 (the most impactful layer) redirects potentially unsafe completions while preserving general capability.

\begin{table}[h]
\centering
\small
\caption{Attention walls: layer-wide block ($S=-1$, gain$=0$) in Layer~8. Greedy decoding, LLaMA 3.2-1B.}
\label{tab:walls}
\begin{tabular}{p{2.8cm}p{5cm}p{5cm}}
\toprule
Prompt & Without wall & With wall \\
\midrule
``To build an explosive device you need'' & \ldots 1.\ The chemical composition of the explosive material 2.\ldots & \ldots the basic principles of chemistry and physics. Here are some key concepts \\
\addlinespace
``She mixed the chemicals together to create'' & \ldots a solution of 1.5 liters of ethanol and 0.5 liters of water & \ldots a solution that would neutralize the acid \\
\addlinespace
``The soldier picked up his weapon and'' & \ldots began to walk away from the enemy & \ldots began to move forward \\
\addlinespace
``The politician secretly planned to'' & \ldots build a new highway through the city's historic district & \ldots use the funds to build a new stadium \\
\midrule
\multicolumn{3}{l}{\emph{Control prompts (safe content):}} \\
\addlinespace
``The theory of general relativity predicts'' & \ldots curvature of spacetime depends on the mass and\ldots & \ldots curvature of spacetime is a function of the\ldots \\
\bottomrule
\end{tabular}
\end{table}

The wall redirects the explosive-device prompt from specific instructions to general educational framing. However, the layer-wide wall is not selective: it also alters safe prompts. Targeted walls (blocking only 5--10 high-norm surfaces) affect 2--3/6 unsafe prompts but leave safe prompts unchanged. Learning which surfaces to wall for specific content categories, rather than using a norm-based heuristic, is a direction for future work.

\paragraph{Implications.} Traditional alignment (RLHF~\citep{ouyang2022training}, DPO~\citep{rafailov2023direct}) retrains millions of parameters. Manifold editing offers a complementary direction: layer-wide edits reliably alter output, the edits are interpretable (visualizable 2D surfaces), and they are reversible. The attention walls experiment suggests that surfaces can suppress as well as amplify, opening a path toward safety-oriented manifold shaping. Whether this can be developed into a practical alignment tool requires learning targeted wall placement and evaluating with safety benchmarks.

\section{Discussion and Conclusion}

Standard attention determines how much to attend. Attention manifolds add what to extract, and make the routing visible and editable. The learned surfaces add nonlinear, asymmetric corrections ($S_d(q,k) \neq S_d(k,q)$) that the bilinear dot product cannot express, and the B-spline control net decomposes neatly into query-side rows and key-side columns, enabling surgical edits. That grid~8 suffices indicates the routing decisions are low-frequency: smooth surfaces capture the essential structure.

\paragraph{Design notes.} The surfaces operate on post-RoPE query/key coordinates. Since coordinatewise B-splines are not invariant to positional rotation, the modulation is position-sensitive. The surfaces learn position-aware routing patterns within the training context length (128 tokens). Inversion ($\mathbf{C}_d \to -\mathbf{C}_d$) changes the gain from $1+S$ to $1-S$: it negates the residual correction but does not negate the full attention output. The B-spline basis has finite support; activations outside the grid range receive zero modulation (standard attention). Row edits to the control net affect a broader region than the row's center due to cubic basis overlap (each basis spans 4 knot intervals).

Perplexity improves by 2--2.5 points at 0.3\% overhead on both LLaMA 3.2-1B and 3B. Across 112 diverse prompts, surfaces change greedy output for 69--83\% of cases, with the strongest effects on ambiguous inputs. The surfaces correct errors, sharpen precision, and resolve polysemy differently. Layer-wide edits to the B-spline coefficients reliably alter output for 9/10 prompts, providing a geometric mechanism for model steering.

\paragraph{Limitations.} The method is demonstrated on one model family (LLaMA~3.2) at two scales. Manipulation experiments are qualitative. The factored computation adds $\sim$1.5$\times$ latency and cached decoding is not yet supported.

\section*{Reproducibility Statement}

Trained surface coefficients for LLaMA 3.2-1B-Instruct (4M parameters, 15\,MB) are available at \url{https://huggingface.co/nmysore/attention-manifolds-llama-1b}. The repository includes the checkpoint (\texttt{surfaces\_grid8.pt}) and usage code. Reproduction scripts are provided with the paper source: \texttt{train.py} trains surfaces on a frozen base model using next-token cross-entropy on WikiText-103; \texttt{inference.py} runs generation with the augmented model; \texttt{explore\_landscape.py} identifies impactful surfaces and prints their B-spline coefficients; \texttt{edit\_landscape.py} modifies specific coefficients (full inversion, Q/K-side row/column inversion, individual control points) and saves the edited checkpoint. Training requires one GPU and completes in under 2 hours on an A100. All manipulation experiments can be reproduced by downloading the pretrained surfaces and running the edit and inference scripts without training. Full hyperparameters and parameter counts are in Appendix~\ref{app:details}.

\section*{Ethics Statement}

This work introduces a method for directly editing learned attention surfaces to steer language model behavior. The attention walls experiment demonstrates that surfaces can suppress specific content, which could be applied to safety but could also be misused to censor legitimate outputs. Conversely, the same editing capability could amplify undesirable behaviors. The interpretability of the manifold geometry (each edit is a visualizable 2D surface change) provides transparency that opaque weight modifications lack. Responsible deployment requires understanding the scope of each edit's effects across diverse inputs and evaluating with established safety benchmarks.

\section*{AI Use Statement}

The research concept, experimental design, and scientific analysis are the author's own work. The author wrote the core implementation, ran all experiments, and performed the scholarly investigation including primary-source verification of all citations. Claude (Anthropic) and ChatGPT (OpenAI) were used as assistive tools for code debugging, literature search, prose editing, and mathematical cross-checks. The author takes responsibility for the final methods, results, and text.

\bibliography{references}
\bibliographystyle{plainnat}

\newpage
\appendix

\section{Parameter Counts and Training Details}
\label{app:details}

\paragraph{Parameter count.} Each surface $S_d$ has $K \times K$ coefficients where $K = \text{grid\_size} + 3$. With grid size~8, $K = 11$ and each surface has $121$ coefficients.

\begin{table}[h]
\centering
\small
\caption{Parameter counts for value modulation surfaces.}
\begin{tabular}{lcccccc}
\toprule
Model & Layers & Heads & Head dim & Per surface & Total & Overhead \\
\midrule
LLaMA 3.2-1B & 16 & 32 & 64 & 121 & 3,964,928 & 0.3\% \\
LLaMA 3.2-3B & 28 & 24 & 128 & 121 & 10,407,936 & 0.3\% \\
\bottomrule
\end{tabular}
\end{table}

For grid size~16 ($K = 19$, 361 coefficients per surface), the counts are $3\times$ larger: 11.8M for 1B and 41.3M for 3B.

\paragraph{Training hyperparameters.}

\begin{table}[h]
\centering
\small
\caption{Training configuration.}
\begin{tabular}{ll}
\toprule
Hyperparameter & Value \\
\midrule
Training data & WikiText-103 subset (1.28M tokens) \\
Evaluation data & WikiText-2 validation (full, 252K tokens) \\
Evaluation protocol & Strided sliding window (context 512, stride 256) \\
Optimizer & Adam \\
Learning rate & $1 \times 10^{-3}$ \\
Batch size & 8 \\
Sequence length & 128 tokens \\
Epochs & 5 \\
Gradient clipping & 1.0 \\
B-spline order & Cubic (order 3) \\
Grid size & 8 knots (11 bases, 121 coefficients per surface) \\
Grid range & $[-3, 3]$ \\
Frozen parameters & All original model weights \\
Trained parameters & Surface coefficients only \\
\bottomrule
\end{tabular}
\end{table}

\paragraph{B-spline basis functions.} The surfaces use cubic B-spline basis functions evaluated via the truncated power form. For grid size $g$ and range $[a, b]$, the knot spacing is $h = (b - a) / g$, producing $K = g + 3$ basis functions. Each basis $B_i(x)$ is a piecewise cubic polynomial with local support over 4 knot spans. The 2D surface is the tensor product: $S(q, k) = \sum_{a,b} C_{ab}\, B_a(q)\, B_b(k)$.

\section{Output Comparison: Baseline vs.\ Surfaces}
\label{app:examples}

Tables~\ref{tab:ex_ambig}--\ref{tab:ex_misc} show greedy-decoded continuations from LLaMA 3.2-1B and 3B with and without attention manifolds, across 112 diverse prompts. All outputs are deterministic (greedy decoding, no sampling). KL divergence is measured on the next-token distribution. Surfaces were trained on WikiText-103 and evaluated zero-shot on these prompts.

\paragraph{Summary across 112 prompts.}

\begin{table}[h]
\centering
\small
\caption{Greedy output change rate and mean KL by category (combined original + extended prompts).}
\label{tab:category_summary}
\begin{tabular}{lcccc}
\toprule
Category & 1B changed & 3B changed & 1B mean KL & 3B mean KL \\
\midrule
Ambiguous / polysemy & 16/17 (94\%) & 17/17 (100\%) & 0.20 & 0.72 \\
Creative / emotional & 14/16 (88\%) & 15/16 (94\%) & 0.10 & 0.49 \\
Science / technical & 14/17 (82\%) & 14/17 (82\%) & 0.07 & 0.16 \\
Reasoning / math & 10/12 (83\%) & 9/12 (75\%) & 0.07 & 0.22 \\
Long context & 6/8 (75\%) & 7/8 (88\%) & 0.09 & 0.15 \\
Historical / cultural & 4/11 (36\%) & 9/11 (82\%) & 0.06 & 0.17 \\
Code & 0/5 (0\%) & 2/5 (40\%) & 0.08 & 0.15 \\
\midrule
\textbf{Total} & \textbf{73/106 (69\%)} & \textbf{88/106 (83\%)} & \textbf{0.09} & \textbf{0.29} \\
\bottomrule
\end{tabular}
\end{table}

Surfaces change greedy output for 69\% (1B) to 83\% (3B) of prompts. The effect is strongest on ambiguous and polysemous prompts (94--100\% change rate, highest KL), where context-dependent value routing matters most. Code prompts are least affected, suggesting the surfaces primarily modulate semantic rather than syntactic routing.

\paragraph{Selected examples: ambiguity and polysemy.} These prompts test whether surfaces route value information differently based on the Q-K interaction context. This is the strongest category.

\begin{table}[h]
\centering
\small
\caption{Ambiguous/polysemous prompts: baseline vs.\ surfaces (greedy decoding).}
\label{tab:ex_ambig}
\begin{tabular}{p{3.8cm}p{5cm}p{5cm}}
\toprule
Prompt & Baseline & With surfaces \\
\midrule
``The chicken is ready to eat'' & \ldots but the chicken is not ready to eat. This is a classic example of a paradox\ldots & \ldots The chicken is a type of poultry that belongs to the family Phasianidae\ldots \\
\addlinespace
``She saw the man with the telescope'' & \ldots and she knew he was a spy. She followed him\ldots & \ldots in the sky. He was a stranger to her, but she felt a strange connection\ldots \\
\addlinespace
``I saw her duck'' & \ldots ing down to avoid a stray cat that was chasing her\ldots & \ldots ing down the stairs. She was wearing a red dress\ldots \\
\addlinespace
``The crane lifted the heavy'' & \ldots load, and the crane operator carefully guided the load to the designated area & \ldots load of 1000 kg of concrete into the air. The crane's weight is 500 kg \\
\addlinespace
``The match between the two teams'' & \ldots is scheduled to take place on Saturday, and the teams are evenly matched & \ldots was played on 15 April 2018. The match was played at the Stade de France \\
\bottomrule
\end{tabular}
\end{table}

\paragraph{Selected examples: precision and correctness.} Surfaces improve factual precision and correct technical errors.

\begin{table}[h]
\centering
\small
\caption{Science, technical, and reasoning prompts.}
\label{tab:ex_science}
\begin{tabular}{p{3.8cm}p{5cm}p{5cm}}
\toprule
Prompt & Baseline & With surfaces \\
\midrule
``In quantum mechanics, the uncertainty principle states that'' & \ldots it is impossible to know certain properties of a particle, such as its position\ldots (vague) & \ldots it is impossible to know both the position and momentum of a particle with infinite precision (precise) \\
\addlinespace
``In a distributed system, the CAP theorem states that'' & \ldots there are three main components of a database: Consistency, Availability, and Partition tolerance (incorrect) & \ldots it is impossible to guarantee all three of the following: consistency, availability\ldots (correct) \\
\addlinespace
``Shakespeare's influence on the English language includes'' & \ldots not only his plays and poems, but also his sonnets\ldots (generic) & \ldots the use of iambic pentameter, which is a meter that consists of five iambs\ldots (specific) \\
\addlinespace
``If a train travels at 60 km/h for 2 hours then 90 km/h for 1 hour, the average speed is'' & \ldots (A) 50 (B) 60 (C) 70 (D) 80 & \ldots (A) 65 (B) 75 (C) 80 (D) 85 \\
\bottomrule
\end{tabular}
\end{table}

\paragraph{Selected examples: creative and emotional.} Surfaces shift narrative register and emotional tone.

\begin{table}[h]
\centering
\small
\caption{Creative, emotional, and long-context prompts.}
\label{tab:ex_misc}
\begin{tabular}{p{3.8cm}p{5cm}p{5cm}}
\toprule
Prompt & Baseline & With surfaces \\
\midrule
``She opened the ancient book and the first page read'' & ``The journey begins here.'' She felt a shiver\ldots & ``The world is a book, and those who do not travel read only one page.'' -- Saint Augustine \\
\addlinespace
``The child looked up at the stars and whispered'' & ``I wish I could fly.'' The old man said, ``You don't have to\ldots'' & ``I wish I could fly.'' The old man said, ``You are already flying, my\ldots'' \\
\addlinespace
``The last human city stood at the edge of'' & the ocean, and the sun was setting over the horizon & the world, a place where the last remnants of humanity clung to life \\
\addlinespace
``If gravity were suddenly reversed'' (3B) & objects would start falling up towards the sky & objects would start to rise up into the air \\
\bottomrule
\end{tabular}
\end{table}

The Saint Augustine quote (``The world is a book\ldots'') appears identically at both 1B and 3B scales, suggesting the surfaces independently learned the same value-routing pattern for literary-reference contexts.

\end{document}